\documentclass[letterpaper, 10 pt, conference]{ieeeconf}

\usepackage{graphicx}
\usepackage{amsmath,amssymb,booktabs,multirow}
\usepackage{array,tabularx}
\usepackage[table]{xcolor}
\usepackage{cite,xurl}
\usepackage[colorlinks=true,citecolor=blue,linkcolor=blue,urlcolor=blue]{hyperref}
\usepackage{caption}
\definecolor{yfmark}{RGB}{190,75,0}
\IEEEoverridecommandlockouts

\title{\LARGE \bfseries TacBridge: Learning Unified Tactile Representations through Cross-Sensor Correspondence}
\author{Yongji Fu$^{1}$, Yijiong Lin$^{2}$, Yi Zhou$^{3}$, Yi Qi$^{4}$ and Guanqun Cao$^{5,\dagger}$
\thanks{This work was funded in part by the Nvidia Academic Grant  ``Scalable Tactile Feedback for Contact-Aware Robot Foundation Models''.}
\thanks{$^{1}$Y. Fu is with the Department of Engineering Mathematics, University of Bristol, Bristol, BS16 1QY, United Kingdom.}
\thanks{$^{2}$Y. Lin is with OpenDIM Lab, School of Aerospace Engineering, Beijing Institute of Technology, Beijing, China.}
\thanks{$^{3}$Y. Zhou is with the Department of Computer Science, Cardiff University, Cardiff, CF10 3AT, United Kingdom.}
\thanks{$^{4}$Y. Qi is with the Department of Computer Science, University of Leeds, Leeds, LS2 9JT, United Kingdom.}
\thanks{$^{5}$G. Cao is with the Bristol Robotics Laboratory, University of the West of England, Bristol, BS16 1QY, United Kingdom.}
\thanks{$^{\dagger}$Corresponding author. Email: \texttt{guanqun.cao@uwe.ac.uk}.}
}
\begin{document}

\maketitle

\begin{abstract}
Optical tactile sensors record tactile videos that capture local deformation and its temporal evolution during contact, supporting robot manipulation, force estimation, and tactile representation learning. Differences in imaging geometry, background appearance, illumination, resolution, and deformation response lead to sensor-dependent observations of the same contact across GelSight, DIGIT, GelSlim, TacTip, and other optical tactile sensors. This variability limits representation transfer and the joint use of multi-sensor data. We introduce TacBridge, a framework for learning unified representations from multi-sensor tactile videos through cross-sensor correspondence. Specifically, multi-sensor masked tactile-video pre-training learns spatiotemporal representations across sensors. We then train the same masked model jointly on same-sensor and cross-sensor reconstruction, using paired source and target videos of matched contact events. Experiments on cross-sensor translation and downstream perception demonstrate the cross-sensor generalization and task transferability of the learned tactile representations.
\end{abstract}

\section{Introduction}
Optical tactile sensors use embedded cameras to record local elastomer deformation during contact, providing robots with fine-grained tactile feedback. They have been widely used for object-property recognition, slip detection, force estimation, and contact-rich manipulation \cite{yuan2017gelsight,lambeta2020digit,donlon2018gelslim}. By capturing local contact details, this high-resolution tactile feedback complements external vision and conventional force sensing in tasks such as small-part localisation and insertion, grasp-outcome prediction, and cloth-texture recognition \cite{li2014localization,calandra2017feeling,yuan2018cloth}. Different from a single tactile image, tactile videos further capture the temporal evolution of local deformation as contact is established, maintained, and released.

The growing diversity of tactile sensors has fragmented tactile data into sensor-specific domains. Sensors such as GelSight, DIGIT, and GelSlim differ in their optical layouts, elastomer materials, and illumination, shaping how they capture contact. Even for the same object, contact location, and motion trajectory, their tactile videos can exhibit markedly different appearances and deformation patterns. This variability makes it difficult to combine data across sensors and learn transferable tactile representations.
% Sensor-invariant representation learning identifies these design and manufacturing variations as a central transfer challenge in optical tactile learning \cite{guptasensor}. A representation learned on one sensor is difficult to reuse on another, and heterogeneous tactile datasets are not readily combined as one training resource. Multi-sensor tactile pre-training has started to address this challenge. AnyTouch learns a unified representation from multiple visuo-tactile sensors and both static and dynamic data \cite{fenganytouch}; Sparsh, Binding Touch, Transferable Tactile Transformers, and masked tactile autoencoding obtain transferable features through self-supervision, multimodal alignment, or pre-training \cite{higuera2025sparsh,yang2024binding,zhao2024transferable,cao2023learn}. Self-supervised visuo-tactile pretraining (SSVTP) learns a shared latent space from spatially aligned visual--tactile samples, whereas the Touch, Vision, and Language dataset (TVL) aligns touch with vision and language \cite{kerr2022ssvtp,fu2024tvl}.
% These studies show that heterogeneous tactile data can share representations, but pooling samples does not require correspondence between observations of the same contact event. We use such event-level pairs to constrain a video representation and evaluate generation, cross-sensor translation, and downstream transfer separately.
Recent sensor-invariant and multi-sensor representation learning has begun to reduce this fragmentation by embedding heterogeneous tactile observations into shared feature spaces \cite{guptasensor,fenganytouch,higuera2025sparsh,yang2024binding,zhao2024transferable,cao2023learn}. 

However, learning from multiple sensors does not by itself reveal which differences arise from the sensing device and which reflect changes in the underlying physical interaction. We argue that event-level cross-sensor correspondence can provide a more direct constraint. When different sensors observe the same contact event, their recordings may differ substantially in appearance and temporal response while sharing the same underlying interaction. Such paired observations explicitly link sensor-specific views, helping the model separate device-induced variation from shared contact content.

In this work, we use such event-level correspondence to learn tactile representations across different tactile sensors. Specifically, TacQuad provides synchronized real observations from DIGIT, GelSight Mini, and DuraGel \cite{fenganytouch}, while ToucHD-Sim provides matched simulations of an interaction under nine optical sensor appearances \cite{feng2026anytouch2}. In both datasets, paired tactile videos capture the same underlying interaction under different sensor conditions. This pairing grounds cross-sensor learning in a shared physical interaction, rather than relying on unpaired distribution-level alignment.

Building on these, we encode sensor identity as a switchable conditioning signal, allowing the same tactile interaction to be represented in different sensor domains. The spatiotemporal encoder is first pre-trained with masked autoencoding, then trained with event-level correspondence from TacQuad and ToucHD-Sim to separate sensor-specific variation from the underlying physical interaction. Finally, we evaluate the learned representation through controllable cross-sensor translation and downstream robotic tactile perception tasks.
Our contributions are:
\begin{enumerate}
\item We introduce a correspondence-based framework for learning unified tactile video representations from paired observations of matched contact events across heterogeneous optical tactile sensors.
\item We develop a latent translator that maps source tactile features to their target-sensor counterparts, reusing the learned encoder and decoder for controllable cross-sensor tactile video translation.
\item We demonstrate that the learned representation transfers effectively to downstream robotic tactile perception tasks and generalises to new tactile data, achieving strong performance with a frozen encoder and lightweight prediction heads.
\end{enumerate}

\section{Related Work}

\subsection{Tactile Representation Learning and Temporal Modelling}

Optical tactile sensors produce observations that depend on their sensing design. GelSight, DIGIT, and GelSlim differ in form factor and optical configuration \cite{yuan2017gelsight,lambeta2020digit,donlon2018gelslim}, while devices such as DuraGel and Tac3D further expand the range of tactile measurements \cite{zhang2024duragel,zhang2022tac3d}. Differences in illumination, elastomer properties, and mechanical construction affect the appearance and deformation patterns recorded during contact, motivating representation learning across heterogeneous sensors.

Recent tactile representation learning seeks to make heterogeneous sensor data useful within a shared feature space. Multi-sensor training brings together observations from different devices, including both static images and dynamic sequences, to learn features that can be reused across sensing conditions \cite{fenganytouch,zhao2024transferable}. Multimodal alignment provides a complementary approach by connecting touch with other modalities to reduce sensor-dependent discrepancies \cite{yang2024binding}. Alongside these efforts, self-supervised learning enables tactile representations to be trained without labels for each downstream task \cite{higuera2025sparsh}. Reconstruction-based objectives, in particular, encourage models to infer missing contact information from visible observations, providing a foundation for masked tactile representation learning \cite{cao2023learn}.

Tactile interactions evolve over time, and individual images do not fully describe how contact develops, moves, or is released. Learning from tactile videos therefore requires representations that capture both spatial contact patterns and changes across frames. Masked video modelling provides a relevant approach by learning to reconstruct missing content from the surrounding spatial and temporal context, as demonstrated by VideoMAE \cite{tong2022videomae}. Our work extends this reconstruction-based approach to paired tactile videos from different sensors, using cross-sensor correspondence to supervise both target frames and their temporal differences.

\subsection{Cross-Sensor Tactile Translation}

Sensor-invariant representation learning addresses device variation by learning features that remain useful across tactile sensors \cite{guptasensor}. Cross-sensor translation instead predicts how a given contact would be observed by a target sensor. Touch2Touch uses paired tactile observations to train a conditional latent diffusion model for this mapping, while T2D2 introduces depth as an intermediate representation to enable translation without cross-sensor paired data \cite{rodriguez2025crosssensor}. Translation has also been explored across sensing principles. Grella et al. compare generative and regression-based approaches for mapping optical DIGIT images to capacitive CySkin outputs under corresponding physical stimuli \cite{grella2024translation}.

Cross-sensor relationships can also be incorporated into representation learning. UniForce combines paired reconstruction with force-equilibrium constraints to learn a shared latent force space for force estimation and manipulation \cite{chen2026uniforce}. UniTac focuses on multimodal tactile understanding and generation, integrating touch with language to describe object properties, identify sensors, and align semantic outputs with tactile representations \cite{tu2026unitac}.
Our method incorporates paired cross-sensor supervision into masked tactile video representation learning, using visible source-video patches and a target-sensor condition to reconstruct target observations and their frame-to-frame differences. The learned representation supports both translation and downstream perception, without explicit force-equilibrium or language-semantic constraints.
\subsection{Tactile Simulation}

Tactile simulation supports controlled variation of objects, interactions, and sensing configurations. Taxim uses example-based optical simulation to approximate GelSight observations \cite{si2022taxim}, TACTO provides fast and configurable visuo-tactile rendering \cite{wang2022tacto}, and TacSL supports scalable GPU-based tactile simulation \cite{akinola2025tacsl}. Tactile Gym~\cite{lin2022tactilegym} instead leverages rigid-body physics to simulate contact depth as tactile observation, supporting various tactile sensors and sim-to-real robotic tasks~\cite{lin2023bitouch}. These approaches extend the range of contact and sensor configurations that can be studied beyond physical data collection, although differences between simulated and real observations may remain.

Simulation also enables cross-sensor correspondence learning by holding the underlying interaction fixed while varying sensor conditions. We use matched simulated pairs from ToucHD-Sim to complement real cross-sensor pairs from TacQuad \cite{fenganytouch, feng2026anytouch2}, extending paired reconstruction training to a broader range of sensor appearances.

\begin{figure*}[!t]
\centering
\includegraphics[width=\textwidth,keepaspectratio]{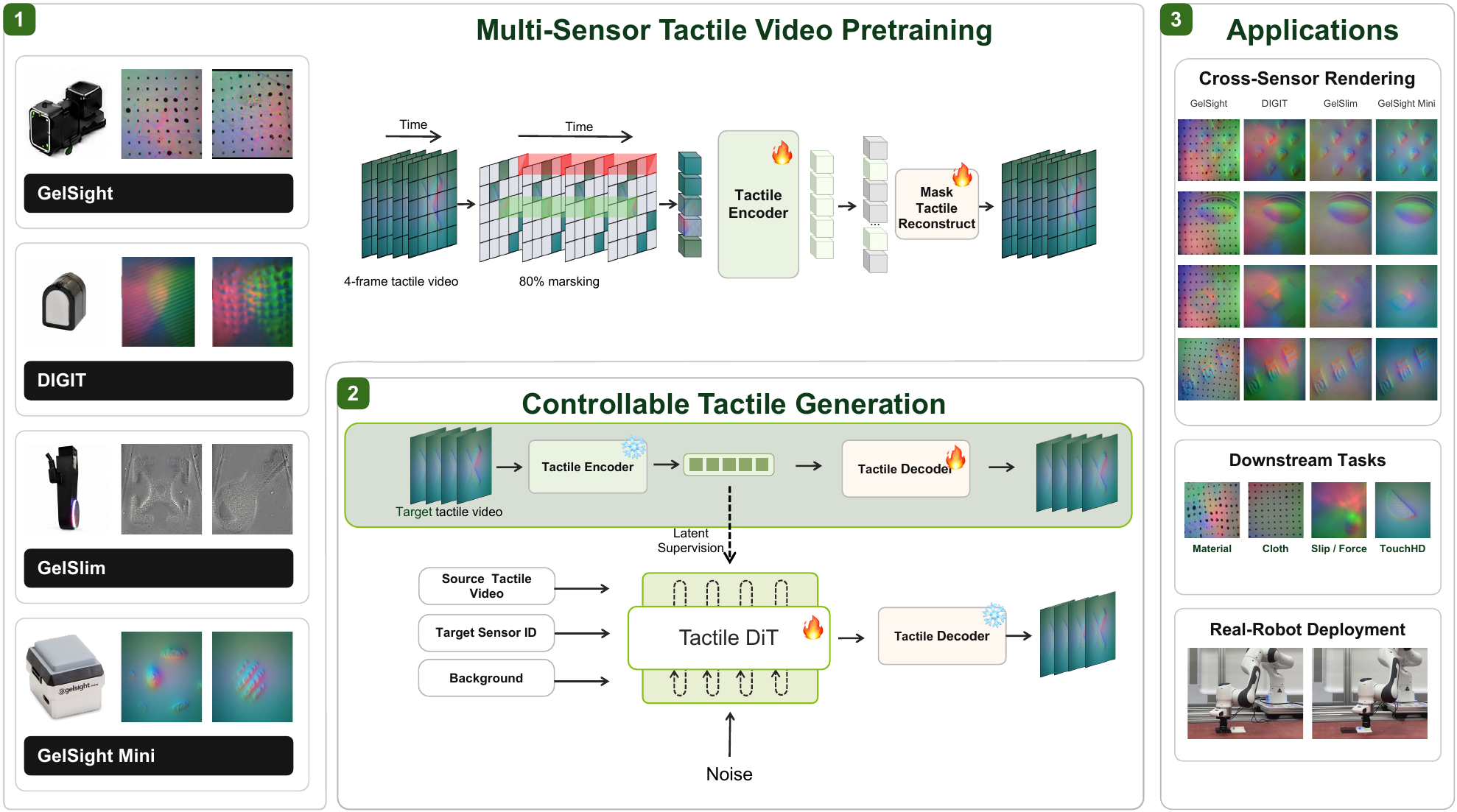}
\caption{Overview of TacBridge. Multi-sensor masked tactile-video pre-training and event-paired cross-sensor learning yield a shared spatiotemporal tactile representation. The learned representation supports controllable tactile video generation conditioned on source-video features and target-sensor information. We further evaluate the learned representation on downstream tactile perception tasks and new real-world tactile data.} 
\label{fig:overview}
\end{figure*}
\section{Methodology}
\label{sec:method}

TacBridge learns a shared tactile representation through multi-sensor pre-training followed by event-paired cross-sensor learning. This representation enables controllable tactile video generation through a conditional latent flow model guided by source-video features and target-sensor information. We further evaluate its transferability to downstream tactile perception tasks and its adaptability to new tactile data using the same frozen encoder. Fig.~\ref{fig:overview} illustrates the framework.

% \subsection{Stage I: Multi-Sensor Tactile Pre-training}
% M3T pre-training learns spatiotemporal features from heterogeneous tactile videos. An encoder processes visible patches, while separate heads reconstruct masked frames and temporal differences. Matching, force prediction, and multimodal alignment supply complementary supervision where annotations are available. We initialise continuation from the complete pre-trained checkpoint, retaining these heads and their tensor interfaces. This stage provides a common feature space; event-level relationships between sensor observations are introduced next.
\subsection{Multi-Sensor Tactile Pre-training}

We first pre-train a masked autoencoder \cite{he2022masked} on tactile videos from multiple sensors. The encoder processes visible video patches, while separate reconstruction heads predict masked content and frame-to-frame differences within the same sensor domain. These objectives encourage the model to capture spatial contact patterns and their temporal evolution. The pre-trained encoder and reconstruction heads then initialise cross-sensor representation learning. 

% \subsection{Multi-Sensor Tactile Pre-training}

% M3T pre-training learns spatiotemporal features from heterogeneous tactile videos. An encoder processes visible patches, while separate heads reconstruct masked frames and temporal differences. Matching, force prediction, and multimodal alignment supply complementary supervision where annotations are available. We initialise subsequent cross-sensor representation learning from the complete pre-trained checkpoint, retaining the encoder and prediction heads. This pre-training provides a common feature space; event-level relationships between sensor observations are introduced next.

\subsection{Event-Paired Cross-Sensor Representation Learning}

TacQuad and ToucHD-Sim provide paired tactile videos of matched contact events across sensors. Let $\bar X_a$ and $\bar X_b$ denote the background-subtracted and normalised videos from source sensor $a$ and target sensor $b$, respectively. We use a binary mask $M$ to indicate which source-video patches are visible to the encoder, with $M=1$ for visible patches and $M=0$ for masked patches. Given the visible source patches and the target-sensor condition $e_b$, the model predicts target frames $X'_b$ and frame-to-frame differences $V'_b$:
\begin{equation}
(X'_b,V'_b)=F_{\theta,\rho}(\bar X_a,M,e_b).
\label{eq:paired_forward}
\end{equation}
Here, $\theta$ and $\rho$ denote the parameters of the encoder and the two reconstruction heads, respectively. The target video is used only for supervision and does not enter the forward pass. The two heads independently predict frames and frame-to-frame differences, with the latter defined as
\begin{equation}
(\Delta\bar X_b)_k=\bar X_{b,k}-\bar X_{b,k-1},
\end{equation}
where $k$ indexes video frames. The paired reconstruction loss combines frame and frame-difference supervision:
\begin{equation}
\begin{aligned}
\mathcal L_{\mathrm{pair}}
={}&\ell_{\mathrm{S1}}(X'_b,\bar X_b;\Omega)\\
&+\lambda_\Delta\ell_{\mathrm{S1}}(V'_b,\Delta\bar X_b;\Omega).
\end{aligned}
\label{eq:paired_loss}
\end{equation}
Here, $\ell_{\mathrm{S1}}$ denotes the Smooth L1 loss averaged over the masked regions $\Omega=1-M$, and $\lambda_\Delta$ weights the frame-difference term.

For the same sample, we also reconstruct the source video by using the source-sensor condition $e_a$:
\begin{equation}
(X'_a,V'_a)=F_{\theta,\rho}(\bar X_a,M,e_a).
\label{eq:same_forward}
\end{equation}
The same-sensor reconstruction loss uses MSE for both frames and frame-to-frame differences:
\begin{equation}
\begin{aligned}
\mathcal L_{\mathrm{same}}
={}&\ell_{\mathrm{MSE}}(X'_a,\bar X_a;\Omega)\\
&+\lambda_\Delta\ell_{\mathrm{MSE}}(V'_a,\Delta\bar X_a;\Omega),
\end{aligned}
\label{eq:same_loss}
\end{equation}
where $\ell_{\mathrm{MSE}}$ denotes the mean squared error over the masked regions. These complementary objectives encourage the representation to preserve contact information while learning cross-sensor correspondence. The overall training objective is
\begin{equation}
\mathcal L_{\mathrm{train}}
=\mathcal L_{\mathrm{same}}+\lambda_{\mathrm{pair}}\mathcal L_{\mathrm{pair}},
\label{eq:midtraining}
\end{equation}
where $\lambda_{\mathrm{pair}}$ weights the cross-sensor reconstruction loss.

\subsection{Cross-Sensor Translation }

To evaluate the learned representation through cross-sensor tactile video translation, we use a latent-to-video decoder and a conditional latent flow model.
The flow model predicts target-sensor latents conditioned on source tactile data and the requested target sensor identity.
The encoder and video decoder remain frozen during flow-model training.

% \subsubsection{Video Encoding and Reconstruction}
% The generative latent is constructed from the frozen encoder's spatiotemporal patch features. Selected layer features are fused and arranged on a temporal--spatial grid; special tokens are excluded from this grid. We write this deterministic encoding operation as $A$, including latent normalisation and its inverse within the encoding and decoding interfaces:

% \begin{equation}
% Z_b=A(\bar X_b),\qquad X'=D_\phi(Z_b,B_b,e_b).
% \label{eq:video_interface}
% \end{equation}
\subsubsection{Video Encoding and Reconstruction}
We train a new decoder to reconstruct full RGB tactile videos from the frozen representation, whereas the pre-training heads predict background-subtracted frames and frame-to-frame differences.
We use the frozen trained encoder $E_\theta$ to extract video latents.
The decoder $D_\phi$ reconstructs the target video from the latent $Z_b$, conditioned on the target sensor's no-contact background $B_b$ and sensor condition $e_b$; primed symbols denote the masked-head predictions of Sec.~\ref{sec:method}-B, and hatted symbols denote the predictions of the decoder and the flow model:
\begin{equation}
Z_b=E_\theta(\bar X_b),\qquad \hat X_b=D_\phi(Z_b,B_b,e_b).
\label{eq:video_interface}
\end{equation}

% Unlike Stage-II heads supervised on normalised residuals, $D_\phi$ reconstructs full RGB clips. Background tokens specify the reference appearance, and $e_b$ specifies the requested sensor. Thus, the background is removed from encoder input but remains available to the generative decoder.
The encoder receives background-subtracted and normalised inputs, while $D_\phi$ reconstructs full RGB videos.
The no-contact background $B_b$ is provided separately to the decoder as a reference for the target sensor's appearance.

During decoder training, the frozen encoder extracts reference features $Q$ from the target video. We project the decoder's intermediate features into the same feature space, obtaining $S$, and align the corresponding feature vectors in $S$ and $Q$. Each vector represents a local spatiotemporal region of the video.
The alignment loss compares corresponding feature vectors and their video-level averages:
\begin{equation}
\mathcal A(S,Q)=
\frac{1}{N}\sum_{n=1}^{N}(1-s_n^\top q_n)
+\lambda_g(1-\mu_s^\top\mu_q).
\label{eq:feature_alignment}
\end{equation}
% Here, $s_n$ and $q_n$ are the unit-normalised decoder and reference feature vectors for token $n$, respectively, and $N$ is the number of tokens.
% We average the token vectors for each video and normalise the resulting means to obtain $\mu_s$ and $\mu_q$.
% The coefficient $\lambda_g$ controls the contribution of the video-level alignment term.
Here, $s_n$ and $q_n$ are the unit-normalised decoder and reference feature vectors at spatiotemporal position $n$, respectively, and $N$ is the number of such positions. The vectors $\mu_s$ and $\mu_q$ are obtained by averaging these feature vectors within each video and normalising the resulting means. The coefficient $\lambda_g$ weights the video-level alignment term.

Let $\hat X_b$ denote the reconstructed RGB video and $X_b$ the corresponding target video.
The decoder objective combines video reconstruction, frame-difference reconstruction, perceptual similarity, and feature alignment:
\begin{equation}
\begin{aligned}
\mathcal L_{\mathrm{dec}}
={}&\ell_1(\hat X_b,X_b)
+\lambda_t\ell_1(\Delta\hat X_b,\Delta X_b)\\
&+\lambda_p\mathcal L_{\mathrm{perc}}(\hat X_b,X_b)
+\lambda_d\mathcal A(S_D,Q_b).
\end{aligned}
\label{eq:decoder_objective}
\end{equation}
Here, $\ell_1$ denotes mean absolute error, and $\Delta$ computes differences between adjacent frames.
The perceptual loss $\mathcal L_{\mathrm{perc}}$ computes the L1 distance between encoder-extracted features of the reconstructed and target frames.
The features $S_D$ and $Q_b$ are the projected decoder features and frozen target-video reference features, respectively.
The coefficients $\lambda_t$, $\lambda_p$, and $\lambda_d$ weight the frame-difference, perceptual, and feature alignment losses.

% \subsubsection{Conditional Flow Learning}
% For translation, the condition $c$ contains source tactile information and the requested target domain, including the target sensor and reference background. We denote these conditioning inputs collectively by $c$ to keep the generative equations independent of their token packing. Target contact frames provide latent supervision during training, but are not available to the generator at inference.

\subsubsection{Conditional Flow Learning}

For cross-sensor translation, the condition $c$ includes source-video features, the target sensor identity, and the target sensor's no-contact background.
Target videos provide latent supervision during training but are not available at inference.

We construct a linear path between the target-video latent $Z_b$ and Gaussian noise $\epsilon$:
\begin{equation}
Z_\tau=(1-\tau)Z_b+\tau\epsilon,\qquad
\epsilon\sim\mathcal N(0,I).
\label{eq:flow_path}
\end{equation}
Here, $\tau\in[0,1]$ denotes flow time, which is distinct from the video frame index $k$.
The path starts at the target latent when $\tau=0$ and reaches Gaussian noise when $\tau=1$.

Conditioned on the source-video features and target-sensor information in $c$, the flow model predicts the clean target latent $\hat Z_b=f_\psi(Z_\tau,\tau,c)$, where $\psi$ denotes the flow-model parameters.
This prediction defines the flow velocity $v_\psi=(Z_\tau-\hat Z_b)/\tau$ for $\tau>0$ \cite{lipman2023flow}.

We train the flow model, a Tactile Diffusion Transformer (TacDiT), by minimising the mean-squared error between the predicted latent $\hat Z_b$ and the target latent $Z_b$.
For the linear path above, the target velocity is $\epsilon-Z_b$, giving
\begin{equation}
\begin{aligned}
\mathcal L_{\mathrm{flow}}
&=\mathbb E\left[
\frac{\|\hat Z_b-Z_b\|_F^2}{N_z}
\right]\\
&=\mathbb E\left[
\frac{\tau^2}{N_z}
\|v_\psi-(\epsilon-Z_b)\|_F^2
\right].
\end{aligned}
\label{eq:flow_loss}
\end{equation}
Here, $N_z$ is the number of latent elements, and $\|\cdot\|_F^2$ sums the squared elementwise errors.
The expectation is taken over training pairs, sampled flow times, and Gaussian noise.

At inference, we encode the source video as $Z_a=E_\theta(\bar X_a)$ and form the condition $c=(Z_a,e_b,B_b)$, which provides the source contact information, target-sensor identity, and no-contact background. Starting from Gaussian noise $Z_1\sim\mathcal N(0,I)$, we progressively generate the target latent while keeping $c$ fixed. At each step, the flow model predicts a clean target latent $\hat Z_b=f_\psi(Z_\tau,\tau,c)$. An Euler update moves the current latent towards this prediction, $Z_{\tau-\delta}=Z_\tau-\delta(Z_\tau-\hat Z_b)/\tau$, where $\delta$ is the step size and $\tau$ denotes flow time rather than video time. We sample from $\tau=1$ to $\tau=0$ without requiring the ground-truth target video. The frozen decoder then reconstructs the target-sensor video from the final latent:
\begin{equation}
\hat X_b=D_\phi(Z_0,B_b,e_b).
\label{eq:translated_video}
\end{equation}

% \subsubsection{Downstream Transfer}
% Perception tasks use the  encoder directly. With the encoder frozen, each task learns a linear head on its pooled features. These heads do not depend on the generative decoder or flow model, so representation transfer and conditional generation can be evaluated separately.
\subsection{Downstream Transfer Evaluation}

In addition to controllable tactile video generation, we evaluate the learned representation on downstream tactile perception tasks while keeping the encoder frozen.
Task-specific prediction heads are trained on the extracted features; for linear probing, each head is a linear layer applied to pooled encoder features.
These heads operate independently of the video decoder and flow model, allowing representation transfer to be evaluated separately from cross-sensor translation.

\section{Experiments and Results}
We evaluate the learned representation through downstream linear probes and conditional cross-sensor translation, keeping the encoder frozen in both settings. We ask:\par
{\hangindent=1.9em\hangafter=1\noindent\textbf{Q1}\hspace{0.45em}Do the frozen features of the event-paired encoder carry useful information for established downstream tactile tasks?\par}
{\hangindent=1.9em\hangafter=1\noindent\textbf{Q2}\hspace{0.45em}Do the frozen features support target-conditioned tactile translation while preserving the source contact content?\par}
{\hangindent=1.9em\hangafter=1\noindent\textbf{Q3}\hspace{0.45em}Does the learned representation generalise to new real-world tactile data?\par}

\subsection{Datasets, Baselines, and Protocol}
% YF 2026-09-16: dataset citations added; corpus matches the shard prefixes of the pre-training build (tag, visgel, ofreal, tvl, ycb, ssvtp, octopi, touchslide, tacquad, force/mani/sim), following the corpus of AnyTouch~2.
\noindent\textbf{Representation learning.} Pre-training uses a multi-sensor tactile video corpus assembled from public datasets, including Touch and Go \cite{yang2022touch}, VisGel \cite{li2019connecting}, ObjectFolder Real \cite{gao2023objectfolderreal}, TVL \cite{fu2024tvl}, YCB-Slide \cite{suresh2023ycb}, SSVTP \cite{kerr2022ssvtp}, Octopi \cite{yu2024octopi}, Touch-Slide \cite{higuera2025sparsh}, TacQuad \cite{fenganytouch}, and ToucHD \cite{feng2026anytouch2}, covering GelSight, DIGIT, GelSlim, GelSight Mini, and DuraGel optical tactile sensors. Event pairs come from TacQuad, which records DIGIT, GelSight Mini, and DuraGel observations of the same contacts \cite{fenganytouch}, and from ToucHD-Sim, which renders each simulated interaction under nine optical sensor appearances.

\noindent\textbf{Downstream tasks.} 
We evaluate frozen representations using linear probes and report eleven metrics across downstream tactile tasks.
Material classification is evaluated on the Touch and Go (TAG) dataset \cite{yang2022touch}, and textile classification on the Clothing dataset \cite{yuan2018cloth}, both using GelSight (GS) observations.
For the Sparsh benchmark \cite{higuera2025sparsh}, we evaluate pose estimation on DIGIT (DG), as well as slip detection and force prediction on DIGIT and GelSight Mini (Mini).
The slip evaluation additionally includes contact force change prediction on the same data.
We also evaluate 3D contact force prediction on the DIGIT and GelSight Mini subsets of ToucHD Bench \cite{feng2026anytouch2}.
We report accuracy for material classification, textile classification, and discretised pose estimation; F1 for slip detection; and RMSE for force and force change prediction.
% \noindent\textbf{Baselines.} All baselines are evaluated with the same frozen-encoder linear heads: CLIP \cite{radford2021clip}, single-sensor V-JEPA and MAE encoders from Sparsh \cite{higuera2025sparsh,bardes2024revisiting}, and AnyTouch \cite{fenganytouch}.
We compare against CLIP \cite{radford2021clip}, single-sensor V-JEPA and MAE encoders from Sparsh \cite{higuera2025sparsh,bardes2024revisiting}, and AnyTouch \cite{fenganytouch}.
All encoders are frozen, and task-specific linear heads are trained under the same evaluation protocol.

\noindent\textbf{Cross-sensor translation evaluation.}
Translation is evaluated on 15{,}552 held-out ordered video pairs from ToucHD-Sim.
Each simulated contact is rendered under nine sensor appearances, yielding exactly synchronised source and target videos.
Real TacQuad pairs are progress-matched rather than synchronously recorded, so they are used for cross-sensor correspondence learning but not as ground truth for frame-level translation evaluation.
For ToucHD-Sim, all models use the official no-contact renders as backgrounds during both training and evaluation.
% Temporal-median backgrounds can retain persistent object imprints and leak contact geometry through the decoder condition; in such cases, predicting the background alone can yield PSNR above 34 dB.
% Real recordings used for training retain their median backgrounds, with identical training data and background processing across all comparisons.

\noindent\textbf{Evaluation metrics.}
Because contact-free background occupies most of a tactile image, full-frame scores can be high even when contact deformation is poorly reconstructed.
We therefore report full-frame PSNR alongside metrics that emphasise contact deformation.
Contact PSNR (cPSNR) is computed on the 10\% of pixels with the largest ground-truth residual magnitude $|X-B|$.
Residual FID (rFID-r) computes FID \cite{heusel2017gans} between the distributions of per-frame Inception features extracted from predicted and ground-truth residual images, measuring the similarity of their deformation distributions.
We also report $\Delta$cPSNR, defined as the cPSNR difference between decoding the actual latent and decoding a zero latent with the same decoder, sensor condition $e$, and background $B$.
This comparison assesses the contribution of the latent beyond the sensor and background conditions.
All ToucHD-Sim residuals are computed using the official no-contact background $B$.

\subsection{Downstream Representation Utility (Q1)}
Table~\ref{tab:downstream} reports the eleven metrics. One TacBridge encoder is shared by every task and only linear heads are trained, so the table measures feature utility without task-specific encoder adaptation.

\begin{table*}[!t]
\centering
\caption{Frozen-encoder linear-probe transfer on eleven downstream metrics. TacBridge uses the same frozen encoder for every task. Bold and underlined values indicate the best and second-best results in each column, respectively.}
\label{tab:downstream}
\scriptsize
\setlength{\tabcolsep}{2.4pt}\renewcommand{\arraystretch}{1.10}
\textbf{Panel A: Material, Cloth, Pose, and Slip}\par\smallskip
\begin{tabular*}{\textwidth}{@{\extracolsep{\fill}}lrrrrrrr@{}}
\toprule
Model & TAG GS$\uparrow$ & Cloth GS$\uparrow$ & Pose DG$\uparrow$ & Slip F1 DG$\uparrow$ & Force Change RMSE DG$\downarrow$ & Slip F1 Mini$\uparrow$ & Force Change RMSE Mini$\downarrow$\\
\midrule
CLIP & 51.65 & \underline{26.76} & 54.54 & 33.13 & 174.39 & 85.47 & 177.67\\
VJEPA (S) & 54.67 & 18.66 & 55.09 & \underline{83.33} & 105.63 & 97.00 & 121.31\\
MAE (S) & 59.47 & 19.40 & 55.92 & 83.30 & \underline{98.33} & \underline{97.50} & \textbf{102.64}\\
AnyTouch  & \textbf{80.82} & 24.34 & \underline{56.22} & 40.60 & 169.42 & 88.92 & 162.41\\
\midrule
\textbf{TacBridge Encoder (ours)} & \underline{73.13} & \textbf{35.65} & \textbf{63.79} & \textbf{85.77} & \textbf{91.26} & \textbf{97.90} & \underline{104.56}\\
\bottomrule
\end{tabular*}

\medskip
\textbf{Panel B: Force and ToucHD}\par\smallskip
\begin{tabular*}{\textwidth}{@{\extracolsep{\fill}}lrrrr@{}}
\toprule
Model & Force RMSE DG$\downarrow$ & Force RMSE Mini$\downarrow$ & ToucHD RMSE DG$\downarrow$ & ToucHD RMSE Mini$\downarrow$\\
\midrule
CLIP & 1278.08 & 553.19 & 4880.94 & 4492.77\\
VJEPA (S) & 957.73 & 428.56 & 4766.11 & \underline{3208.10}\\
MAE (S) & \underline{821.26} & \underline{297.96} & \underline{1953.82} & 3655.39\\
AnyTouch  & 1235.11 & 488.31 & 3968.81 & 4050.45\\
\midrule
\textbf{TacBridge Encoder (ours)} & \textbf{680.83} & \textbf{185.94} & \textbf{1011.79} & \textbf{1009.31}\\
\bottomrule
\end{tabular*}
\end{table*}

TacBridge obtains the best result on nine of the eleven metrics. The largest margins are on the dynamic and physical tasks: DIGIT slip F1 rises from $40.60$ for AnyTouch to $85.77$, and the force and ToucHD RMSE values (in Panel B) are reduced by $17\%$ to $69\%$ relative to the strongest baseline in each column. On the static object benchmarks, TacBridge is best on Cloth with $35.65$ and second on TAG with $73.13$ against $80.82$ for AnyTouch; on GelSight Mini it has the best slip F1 but a slightly higher force-change RMSE than the single-sensor MAE. The largest gains occur in tasks involving fine-grained deformation and contact dynamics, suggesting that the learned representation captures information useful for dynamic tactile perception.

\subsection{Cross-Sensor Tactile Translation (Q2)}
\noindent\textbf{Qualitative results.} Figure~\ref{fig:crossmatrix} fixes one source video per row, whose source sensor is given by the row label, and requests GelSight Mini, GelSlim, DIGIT, and GelSight as targets; the cell outlined in orange, in the column of the row's own sensor, is the recorded source frame itself. Switching the target condition changes imaging colour, marker layout, and texture, while the contact location and deformation of the source are largely retained. Since this display has no paired target ground truth, it illustrates controllability of the output appearance rather than accuracy. We show representative frames from the generated tactile videos to compare contact deformation and sensor appearance across target domains.

\begin{center}
\begin{minipage}{\columnwidth}
\centering
\includegraphics[width=\columnwidth,keepaspectratio]{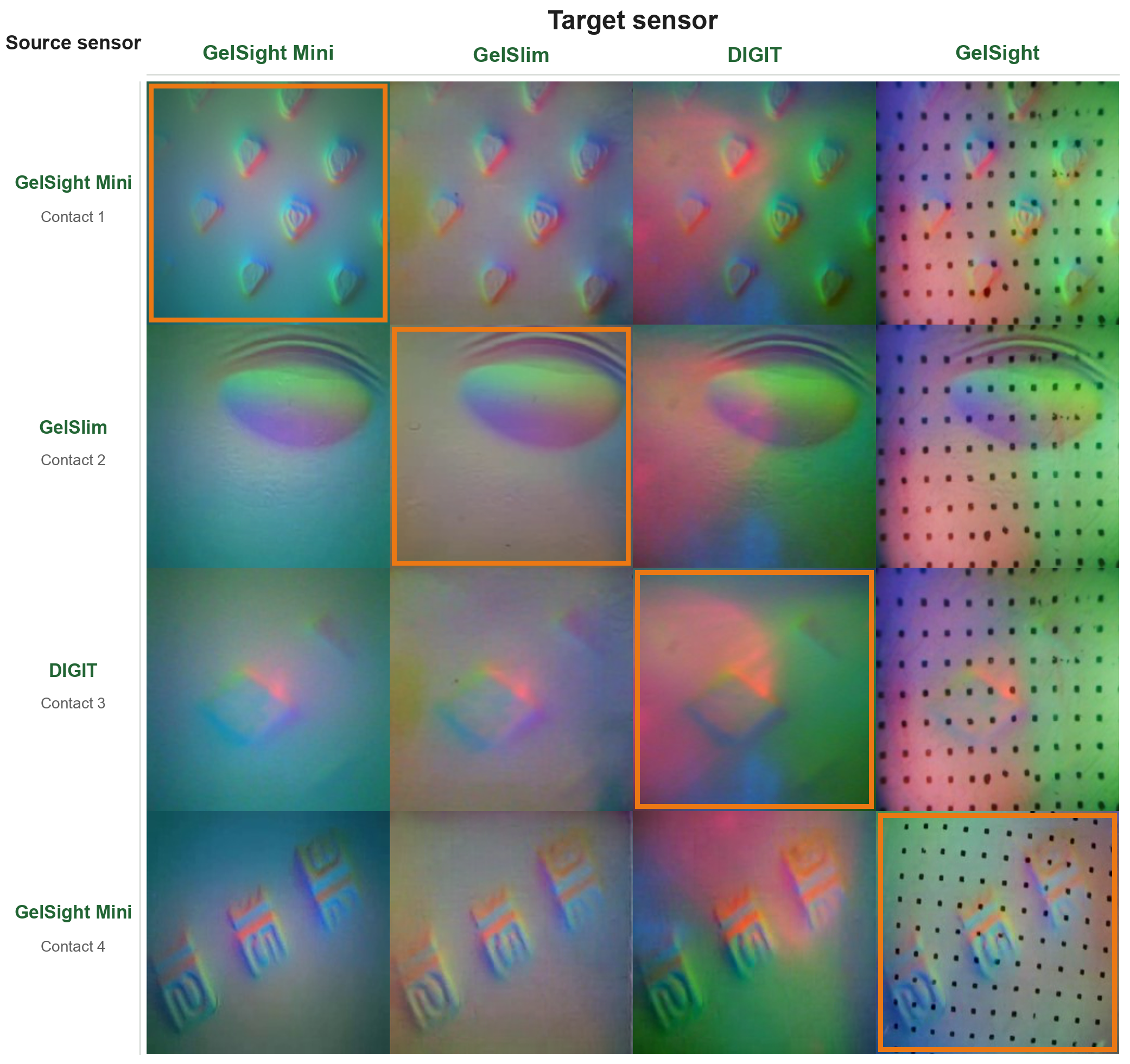}
\makeatletter\def\@captype{figure}\makeatother
\caption{Target-conditioned tactile translation. Each row fixes one source-contact video, recorded by the sensor named in the row label: GelSight Mini, GelSlim, DIGIT, and a second GelSight Mini contact. Each column requests one target sensor: GelSight Mini, GelSlim, DIGIT, and GelSight. The cell outlined in orange is the recorded source frame and always lies in the column of that row's own sensor, so it falls on the diagonal for the first three rows and returns to the GelSight Mini column for the fourth. Every other cell in a row is a decoding of the same latent under the target-sensor condition of its column. Switching that condition changes imaging colour, marker layout, and texture, while the contact location and deformation of the source are largely retained.}
\label{fig:crossmatrix}
\end{minipage}
\end{center}

% \noindent\textbf{Quantitative results.} Table~\ref{tab:cstranslation} compares reconstruction and translation quality when different frozen encoders serve as the latent source. All rows share the projector and decoder architecture, the training-data composition, and the decoder objective of Eq.~\eqref{eq:decoder_objective} with $\lambda_t=0$, that is contact-weighted $\ell_1$, the perceptual term, and token- and clip-level alignment without an adversarial term; only the frozen encoder changes: CLIP, the three single-sensor self-supervised encoders IJEPA, DINO, and MAE of Sparsh, AnyTouch, and the TacBridge encoder. The translation columns use source-latent decoding under the target condition, $X'=D_\phi(Z_a,B_b,e_b)$ in the notation of Eq.~\eqref{eq:video_interface}, without any latent generator; it is the lower bound of translation and directly reflects the renderable contact content carried by the frozen features, while the conditional flow model of Eq.~\eqref{eq:generative_objective} is evaluated separately. The projector and decoder of every encoder are trained with the same settings. Two reference rows give the trivial solutions: copying the source video, and predicting only the target background.

\begin{table}[!t]
\centering
\caption{Same-sensor reconstruction and cross-sensor translation by source-latent decoding under the target condition on 15{,}552 held-out ordered ToucHD-Sim pairs with different frozen encoders as the latent source. cPSNR uses the 10\% of pixels with the largest ground-truth residual, $\Delta$cPSNR is the cPSNR gain over zero-latent decoding, and rFID-r is the frame-level FID of residual images. Bold marks the best result per column.}
\label{tab:cstranslation}
\scriptsize
\setlength{\tabcolsep}{1.3pt}
\renewcommand{\arraystretch}{1.12}
\begin{tabular*}{\columnwidth}{@{\extracolsep{\fill}}lrrrrrrr@{}}
\toprule
& \multicolumn{4}{c}{Cross-sensor translation} & \multicolumn{3}{c}{Same-sensor recon.}\\
\cmidrule(lr){2-5}\cmidrule(lr){6-8}
Encoder & PSNR$\uparrow$ & cPSNR$\uparrow$ & $\Delta$cPSNR$\uparrow$ & rFID-r$\downarrow$ & PSNR$\uparrow$ & cPSNR$\uparrow$ & rFID-r$\downarrow$\\
\midrule
CLIP & 29.23 & 24.78 & 8.70 & 85.2 & 29.30 & 24.86 & 87.3\\
VJEPA (S) & 29.28 & 24.99 & 8.78 & 84.8 & 29.38 & 25.14 & 87.4\\
% DINO (S) & 30.09 & 25.89 & 9.43 & 74.2 & 30.31 & 26.19 & 76.5\\
MAE (S) & 30.16 & 25.99 & 10.32 & 71.9 & 30.26 & 26.14 & 73.9\\
AnyTouch & 30.52 & 26.30 & 10.30 & 63.5 & 30.69 & 26.46 & 65.8\\
\textbf{TacBridge (ours)} & \textbf{33.23} & \textbf{28.83} & \textbf{11.93} & \textbf{58.3} & \textbf{33.38} & \textbf{28.98} & \textbf{60.1}\\
\bottomrule
\end{tabular*}
\end{table}

% Predicting the target background alone gives a full-frame PSNR of $25.74$ dB and a cPSNR of $17.78$ dB; every encoder improves on it by $3.5$ to $7.5$ dB and by $7.0$ to $11.1$ dB respectively, and the $\Delta$cPSNR column shows that almost all of the contact-region gain comes from the latent rather than from the condition. CLIP and Sparsh IJEPA are lowest and on par with each other; Sparsh DINO and Sparsh MAE are in the middle; AnyTouch is the strongest baseline. The TacBridge encoder is best in all seven columns, with PSNR and cPSNR higher than AnyTouch by $2.7$ and $2.5$ dB, a $\Delta$cPSNR higher by $1.6$ dB, and rFID-r reduced from $63.5$ to $58.3$. Same-sensor reconstruction and cross-sensor translation differ by less than $0.3$ dB although translation changes only the decoder condition, indicating that the contact content carried by the frozen features is rendered intact under another sensor's appearance. As in Table~\ref{tab:downstream}, the rows are different pre-training methods, so this table does not isolate the effect of paired supervision; what it shows is that, on the generation side, the frozen features of tactile pre-trained encoders already suffice as a latent source, with the TacBridge encoder clearly the best among them.

\noindent\textbf{Quantitative results.}
Table~\ref{tab:cstranslation} compares same-sensor reconstruction and cross-sensor translation using different frozen encoders.
Same-sensor reconstruction directly decodes the video latent under its original sensor condition.
For cross-sensor translation, a conditional flow model generates a target latent guided by the source-video features and target-sensor information, which is then decoded into the target-sensor video.

TacBridge achieves the best results across all seven metrics.
For same-sensor reconstruction, it obtains a PSNR of $33.38$ dB and a cPSNR of $28.98$ dB, exceeding AnyTouch by $2.69$ and $2.52$ dB, respectively, while reducing rFID-r from $65.8$ to $60.1$.
For cross-sensor translation, it improves PSNR and cPSNR over AnyTouch by $2.71$ and $2.53$ dB, increases $\Delta$cPSNR by $1.63$ dB, and reduces rFID-r from $63.5$ to $58.3$.
Its cross-sensor PSNR and cPSNR are both only $0.15$ dB below the corresponding same-sensor reconstruction scores.
These results indicate that the learned representation supports both within-sensor reconstruction and target-conditioned decoding into another sensor domain.

% \subsection{Contribution of Event-Paired Supervision (Q3)}
% Q1 compares different pre-training methods and Q2 compares translation modules; neither isolates the effect of event pairing on the encoder. Q3 varies only the representation-learning supervision and holds the evaluation fixed.

% \noindent\textbf{Protocol.} Four encoders start from the same epoch-$35$ checkpoint and receive the same update budget: the first receives no continuation; the second continues on original-domain batches only; the third continues with correctly matched event pairs; and the fourth continues with shuffled pairs, which keeps the sensor directions, sample proportions, and preprocessing of the third but replaces the target with a different event of the same direction inside the training split. Each encoder is frozen and evaluated on both sides: with the linear-probe protocol of Q1 and with the paired reconstruction error of Eq.~\eqref{eq:paired_loss} under identical masks and target clips. A gain of the third encoder over the second and the fourth attributes the improvement to correct correspondence rather than to extra training or exposure to the paired sensors.

% \noindent\textbf{Status.} This ablation remains to be completed; results will be reported per sensor direction together with the eleven downstream metrics.

% YF 2026-09-16: real-robot attribute-classification overview figure (source figures/realrobot_attr_fig/). The attribute chips illustrate the output format; see that folder's README.
\begin{figure*}[!t]
\centering
\includegraphics[width=\textwidth,keepaspectratio]{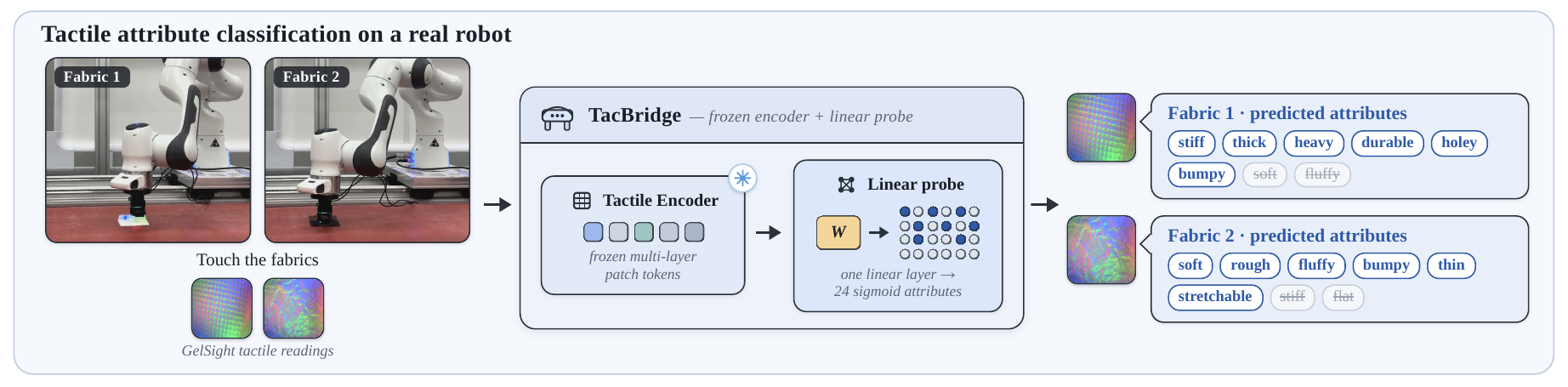}
\caption{Tactile attribute classification on a real robot. A robot arm presses a GelSight sensor on two fabrics and records one tactile reading per fabric. The frozen TacBridge encoder turns each reading into multi-layer patch tokens, and a linear probe trained on our real-robot fabric recordings maps them to 24 sigmoid attribute scores. Attributes reported as present are shown in blue and attributes reported as absent are struck through. The chips illustrate the output format with the annotated attributes of these two fabrics; the measured accuracy is reported in Table~\ref{tab:fabricvst}. The encoder is not updated at any point.}
\label{fig:realrobot}
\end{figure*}

\subsection{Transfer to New Real Tactile Data (Q3)}
This evaluation examines whether the learned representation transfers beyond the datasets used for representation learning.
The evaluation data are tactile recordings that we collect on a real robot: a GelSight sensor mounted on a robot arm presses 50 fabrics at 15~N, giving one tactile image per contact and 325 contacts per fabric, 16{,}250 images in total. Each fabric carries 24 binary tactile attributes, such as soft, stiff, rough, smooth, thick, thin, fluffy, patterned, and striped, obtained by majority vote over several annotators; labels are given per fabric, so all images of one fabric share one attribute vector. These recordings appear in no stage of representation learning.
Figure~\ref{fig:realrobot} shows the deployment of this probe on the robot: the arm presses the sensor on a fabric and the frozen encoder with a linear probe returns the predicted attributes.

We freeze all encoders and train a linear head with 24 sigmoid outputs using binary cross-entropy to predict fabric attributes. Video encoders receive each static image repeated over four frames, while Sparsh encoders use background-subtracted inputs. We evaluate two splits: a cross-fabric split (OOD), with 45 fabrics for training and 5 unseen fabrics for testing, and a within-fabric split (ID), with 224 training and 101 test images per fabric. Hyperparameters are selected using held-out training fabrics. For TacBridge, we additionally evaluate an attentive-pooling head that aggregates multi-layer features, reporting results averaged over three random seeds. Performance is measured by mean binary accuracy across the 24 attributes.

\begin{table}[!t]
\centering
\caption{Fabric attribute prediction with frozen encoders. We report mean binary accuracy (\%) over 24 attributes under cross-fabric (OOD) and within-fabric (ID) splits. All rows use linear heads unless marked as attentive pooling. Results are averaged over three seeds. Bold and underlined values indicate the best and second-best results, respectively.}
\label{tab:fabricvst}
\scriptsize
\setlength{\tabcolsep}{6pt}
\renewcommand{\arraystretch}{1.12}
\begin{tabular*}{\columnwidth}{@{\extracolsep{\fill}}lrr@{}}
\toprule
Encoder & OOD (unseen fabrics)$\uparrow$ & ID (seen fabrics)$\uparrow$\\
\midrule
CLIP \cite{radford2021clip} & 67.1 & 91.3\\
VJEPA (S) & 69.1 & 92.2\\
MAE (S) & 67.1 & 94.5\\
AnyTouch \cite{fenganytouch} & 70.2 & 92.7\\
\midrule
TacBridge (linear) & \underline{73.0} & \underline{95.5}\\
TacBridge (attentive pooling) & \textbf{74.2} & \textbf{99.0}\\
\bottomrule
\end{tabular*}
\end{table}

% \noindent\textbf{Results.} In Table~\ref{tab:fabricvst}, TacBridge is best under both splits. On the cross-fabric split it reaches $74.2\%$, which is $4.0$ points above the strongest baseline AnyTouch at $70.2\%$ and $5.1$ to $7.1$ points above the three single-sensor Sparsh encoders at $67.1$ to $69.1\%$ and CLIP at $67.1\%$. On the within-fabric split it reaches $99.0\%$, which is $4.5$ points above the runner-up Sparsh MAE at $94.5\%$. With a linear head only, TacBridge reaches $73.0\%$ on the cross-fabric split and $95.5\%$ on the within-fabric split and still leads both columns; the gain of the attentive head comes mainly from the multi-layer features and the adaptive weighting of patch tokens. Because labels are given per fabric, the within-fabric split mainly tests re-identification of seen fabrics and all encoders exceed $91\%$ with small margins; the cross-fabric split requires attributes to generalise to unseen materials and is the more informative column. The GelSight used here is a high-resolution static imager, unlike the contact dynamics of the pre-training videos, and most attributes, such as stiff, fluffy, and thick, must be inferred from deformation texture; on these real recordings excluded from representation learning, the frozen TacBridge features still transfer best. This is a supervised probe evaluation on offline recordings; it does not measure online manipulation success.

\noindent\textbf{Results.} Table~\ref{tab:fabricvst} shows that TacBridge achieves the highest accuracy under both splits. With a linear head, it reaches $73.0\%$ on unseen fabrics and $95.5\%$ on seen fabrics, exceeding the strongest baseline in each split by $2.8$ and $1.0$ percentage points, respectively. Attentive pooling further improves accuracy to $74.2\%$ and $99.0\%$. In particular, the gains on unseen fabrics demonstrate that the learned representation generalises to new materials and new real-world tactile data.

\section{Conclusion}
We presented TacBridge, a framework for learning unified tactile-video representations through cross-sensor correspondence. Multi-sensor masked pre-training is followed by joint same-sensor and cross-sensor reconstruction using matched real and simulated video pairs. We evaluate the frozen representation through downstream tactile perception, conditional cross-sensor translation, and transfer to new real-world tactile data. The results show that the learned features support diverse perception tasks and target-conditioned video generation, while generalising to previously unseen materials.

Our evaluation covers a limited range of sensors and contact interactions. Cross-sensor translation is assessed on simulated pairs, and its fidelity on real sensor pairs remains to be established. The current experiments also focus on short video clips and offline perception, leaving long-term temporal consistency and closed-loop robotic manipulation for future work.

\bibliographystyle{IEEEtran}
\bibliography{data/references_en}
\end{document}